\documentclass[10pt,journal,compsoc]{IEEEtran}
\usepackage{amsmath}
\usepackage{caption}
\usepackage{array}
\usepackage[linesnumbered, boxed]{algorithm2e}

\ifCLASSOPTIONcompsoc
  \usepackage[nocompress]{cite}
\else
  \usepackage{cite}
\fi
\ifCLASSINFOpdf
   \usepackage[pdftex]{graphicx}
\else
\fi
\ifCLASSINFOpdf

\else

\fi

\begin{document}

\title{Abnormal Event Detection and Location for Dense Crowds using Repulsive Forces and Sparse Reconstruction}

\author{Pei~Lv,
        Shunhua~Liu,
        Mingliang~Xu and
        Bing~Zhou% <-this % stops a space
\IEEEcompsocitemizethanks{

 \IEEEcompsocthanksitem Pei Lv, Shunhua Liu,  Mingliang Xu and Bing Zhou are with Center for Interdisciplinary Information Science Research, ZhengZhou University, 450000. \{iexumingliang, ielvpei, iebzhou\}@zzu.edu.cn; .
 %\IEEEcompsocthanksitem Meng Wang is with Center for Shared Experimental Education, Sun Yat-Sen University (SYSU), 510275. nielin@mail.sysu.edu.cn.
 }
\thanks{Manuscript received December 26, 2016. This work was supported in part by the National Natural Science Foundation of China under Project 61379079, 61472370, 61502433, 61672469 and in part by China Postdoctoral Science Foundation under Project 2015M582203, 2016T90680.}}

\markboth{IEEE TRANSACTIONS ON IMAGE PROCESSING, }%
{Shell \MakeLowercase{\textit{et al.}}: Bare Demo of IEEEtran.cls for Computer Society Journals}

\IEEEtitleabstractindextext{%
\begin{abstract}
This paper proposes a method based on repulsive forces and sparse reconstruction for the detection and location of abnormal events in crowded scenes. In order to avoid the challenging problem of accurately tracking each specific individual in a dense or complex scene, we divide each frame of the surveillance video into a fixed number of grids and select a single representative point in each grid as the individual to track. The repulsive force model, which can accurately reflect interactive behaviors of crowds, is used to calculate the interactive forces between grid particles in crowded scenes and to construct a force flow matrix using these discrete forces from a fixed number of continuous frames. The force flow matrix, which contains spatial and temporal information, is adopted to train a group of visual dictionaries by sparse coding. To further improve the detection efficiency and avoid concept drift, we propose a fully unsupervised global and local dynamic updating algorithm, based on sparse reconstruction and a group of word pools. For anomaly location, since our method is based on a fixed grid, we can judge whether anomalies occur in a region intuitively according to the reconstruction error of the corresponding visual words. We experimentally verify the proposed method using the UMN dataset, the UCSD dataset and the Web dataset separately. The results indicate that our method can not only detect abnormal events accurately, but can also pinpoint the location of anomalies.
\end{abstract}

}

\maketitle

\IEEEdisplaynontitleabstractindextext

\IEEEpeerreviewmaketitle

\IEEEraisesectionheading{\section{Introduction}\label{sec:introduction}}

\IEEEPARstart{I}{n} recent times, there have been an increasing number of terrorist attacks, crowd stampedes and other public safety events. Detection and location of abnormal crowd events is the foundation of monitoring, analysis and early warnings for crowd movement, and has become one of the most urgent problems in intelligent surveillance. However, this problem has the following challenges: (1) Severe occlusion occurs frequently in crowds and the appearance of many individuals often deform when they are moving. Thus, it is very difficult to analyze crowd behaviors by tracking each specific individual. (2) Crowd behaviors are extremely complex and emergencies often happen without obvious warning signs, so identification of new and effective crowd movement features is one of the main goals for researchers. (3) Each individual in a crowd is not only affected by his/her surroundings and other individuals, but also his/her desired destination. However, previous methods have not paid sufficient attention to this fact.

In order to solve these problems, researchers have attempted to model the detection and location of abnormal crowd behaviors in two stages, event representation~\cite{zhao2011online,kwon2012unified,tamrakar2012evaluation,jiang2011anomalous,calderara2011detecting} and anomaly judgment~\cite{brostow2006unsupervised,tu2008unified,ali2007lagrangian,wu2010chaotic,scovanner2009learning}. For crowd event representation, the most important component is feature selection. However, the features adopted by current methods, such as color, texture, gradient, object silhouette, spatial-temporal trajectory etc., cannot efficiently deal with the challenging problems described above. As pointed out in~\cite{lu2013abnormal}, abnormal event judgment is not a typical classification problem due to the difficulty in listing all possible negative samples. Research in this area commonly uses a training video to firstly learn normal crowd behaviors, and then uses this knowledge to detect events that deviate from the normal representation. In this paper, this technique is also utilized involving new features.

Crowd Event Representation. The proposed method uses particles tracked within fixed grids to model the underlying events for a crowd. The DeepFlow method proposed in~\cite{weinzaepfel2013deepflow} is used as our initial input, which is an improved optical flow method and can handle severe occlusion and large-scale movement. After obtaining the dense optical flow of the moving crowd, we further extract sparse particles from the image grids using a novel filtering strategy. In contrast with the social force model~\cite{helbing1995social,mehran2009abnormal} or the potential energy model~\cite{wu2010chaotic,scovanner2009learning}, our paper adopts a data-driven statistical-based repulsive force model based on these particles in order to describe the interaction between different individuals in the crowd. According to the interaction energy computed directly by this model, the force that drives individuals to avoid collision can be easily inferred. Therefore, this driving force can be used instead of the social force to construct our underlying features. Since it is still very difficult to distinguish between normal and abnormal behaviors using only these discrete force values~\cite{mehran2009abnormal}, we construct a force flow matrix containing spatial and temporal information to model changes in virtual forces among crowds over time.

Abnormal Event Detection and Location. Previous studies~\cite{castellano2009statistical,chate2008collective} have proven that normal crowd events usually have strongly-related features and abnormal events have large deviations from normal events. Based on this fact, abnormal event detection is usually regarded as an event reconstruction process. Visual words are extracted from normal crowd events to firstly build a corresponding group dictionary model. When a new crowd event occurs, existing visual words are used by the model to reconstruct the event. If the reconstruction error is larger than a pre-defined threshold, this event is judged to be an abnormal behavior. Otherwise, it is a normal event. Since our visual words are built on repulsive forces that are extracted from fixed image grids, once an abnormal event is detected, its position is also located. One of the most important advantages of the proposed method is that since the reconstruction efficiency of the model will seriously degrade when the number of word entries in its dictionaries increase, a fully unsupervised dynamic updating algorithm based on sparse reconstruction and a group of word pools is proposed. Updating the dictionary is divided into two aspects: global group dictionary updates and local single dictionary updates, which are decided by the current reconstruction ability of the proposed method.

The main contributions of our work are as follows:

\begin{itemize}
\item A repulsive force is introduced to construct an underlying feature named after the force flow matrix to model the crowd movement. This new feature can not only describe crowds of different velocities and densities, but also can predict future trends in crowd movement, which have not been studied by previous anomaly detection methods.
\item A group dictionary model built on a repulsive force flow matrix is introduced to detect and locate abnormal events in a crowd. This model can achieve higher accuracy and faster detection speed than using only a single dictionary.
\item In order to solve the degradation problem and concept drift of dictionaries, we propose an unsupervised learning and updating algorithm for the group dictionary model, which is based on sparse reconstruction and a group of word pools.
\end{itemize}

%2相关工作
\section{Related works}
Intelligent surveillance has a wide range of applications, including person counting~\cite{cong2009flow}, object tracking~\cite{avidan2007ensemble}, action recognition~\cite{yuan2009discriminative}, pedestrian detection~\cite{dalal2005histograms} and traffic surveillance~\cite{wang2009unsupervised}. Crowd anomaly detection, which is one of the most important problems in intelligent surveillance, has been receiving more and more attention recently. After several years of development, the mainstream technique for abnormal event detection in crowds has evolved from rule-based methods~\cite{anderson2009linguistic,nasution2007intelligent,ivanov1999recognition} to statistic-based methods~\cite{prasad2007machine}. These methods usually contain two important aspects: (1) feature representation and extraction of crowd movements; (2) modeling and detection of abnormal crowd behaviors.
%2.1
\subsection{Feature representation and extraction of crowd movement}
The most important task to describe crowd behaviors is to extract powerful features. At present, some local features, such as histograms of optical flow~\cite{adam2008robust,zhu2014sparse,cong2013abnormal}, spatial-temporal motion patterns~\cite{kratz2009anomaly}, mixtures of dynamic textures~\cite{mahadevan2010anomaly} and salient features~\cite{itti2005principled} have been widely adopted. As well as these local features, some researchers have also used global features to represent crowd behaviors. For example, Mehran et al. ~\cite{mehran2009abnormal} applied a simplified social force model to describe crowd behaviors by estimating the interaction force between individuals. Mehran et al. ~\cite{mehran2010streakline} introduced streaklines integrated with a particle advection scheme that can incorporate the spatial changes in the particle flow. Wu et al. ~\cite{wu2010chaotic} leveraged chaotic invariance to analyze events in both coherent and incoherent scenes. Xu et al. ~\cite{xu2015learning} proposed an Appearance and Motion DeepNet (AMDN) method which utilizes deep neural networks to automatically learn feature representations. In ~\cite{cui2011abnormal}, a potential interaction energy based on linear trajectory avoidance was proposed to represent the behavioral state of subjects.

For abnormal crowd event detection, the selected feature is critical to the final result and also determines which detection method will be used. In this work, the repulsive force is selected to construct the underlying feature, which is named the force flow matrix. This type of feature has the advantages of using both existing local and global features. Moreover, it can describe future crowd movement trends and be used with different crowds of various velocities and densities.
%2.2
\subsection{Abnormal event detection}
Abnormal event detection methods can be divided into different classes based on different perspectives. Some researchers have proposed cluster-based abnormal event detection~\cite{fu2005similarity,calderara2007detection,javan2013online}, which groups crowd video clips into the same class based on similar structural and semantic info and uses a statistical model to describe this class. If a new example is distant from the centers of existing normal events, it will be regarded as an abnormal case.

Another type of similar statistical method is based on dynamic Bayes networks, such as the Hidden Markov Model (HMM). In HMM, the hidden state and the transition matrix are learned from normal crowd movement and abnormal behavior can then be detected by computing the probability when a new example is involved with trained HMMs~\cite{kratz2009anomaly}. When the computed probability is lower than a pre-defined threshold, the crowd behavior is judged to be an anomaly. A topic model, such as LDA~\cite{mehran2009abnormal}, HDP ~\cite{pruteanu2008infinite,hu2009abnormal,wang2009unsupervised} and PLSA ~\cite{varadarajan2009topic,jager2008weakly}, has been recently applied by researchers to detect crowd abnormal behaviors. The documents in the topic model are used to represent video clips and different regions in these clips are treated as visual words. Abnormal events can be detected by computing the correlation between the visual words that are extracted from the new video clips and those in the existing documents.
\begin{figure*}[t]
\begin{centering}
 \centering
  \includegraphics[width=2.0\columnwidth]{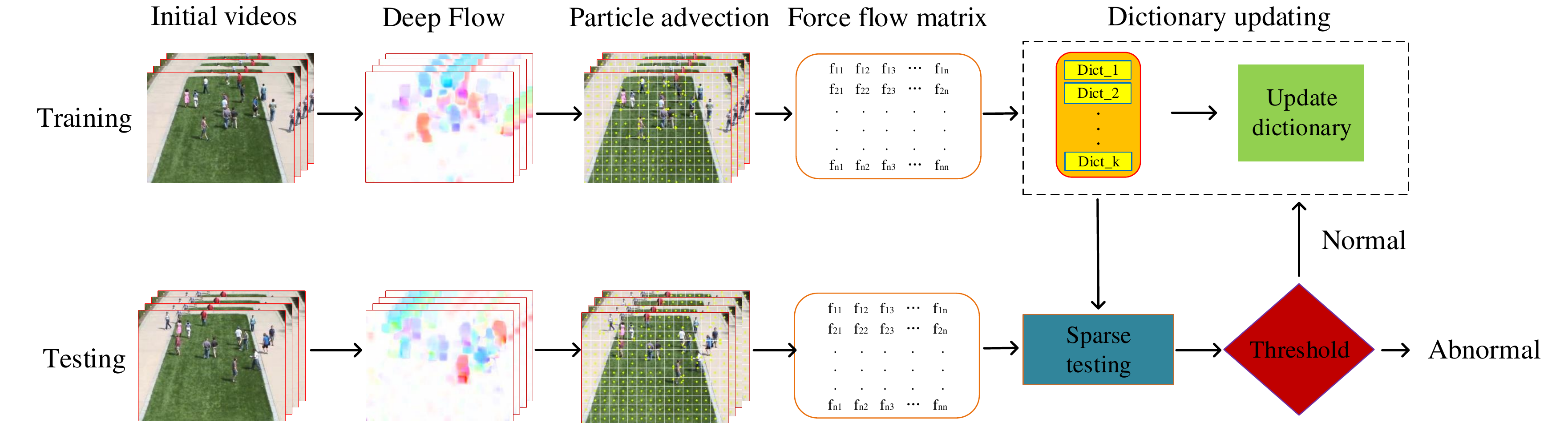}
  \centering
  \caption{System overview of our method. Using sparse particle tracking and a repulsive force model, we firstly construct our underlying features that are named after the force flow matrix. Then we employ sparse coding to train a group of dictionaries which are used for event detection. Finally, to maintain the expression ability of these dictionaries, we propose a novel updating strategy.}
  \label{fig:1}
\end{centering}
\end{figure*}

In both statistics-based models and topic-based models, high-dimensional features are often used to represent crowd events, which will cause the amount of training data to increase exponentially. Even worse, in real scenes, it is impossible to provide enough data to train a complete model that includes all possible crowd scenes. Therefore, some researchers have started using sparse coding to solve this problem~\cite{zhao2011online,zhu2014sparse,cong2011sparse}. The underlying assumption is that normal events can be reconstructed by sparse coefficients with a small error, but for abnormal events, the reconstruction coefficients will be dense and a large error will be generated. Current sparse coding methods for abnormal event detection have overcome the high-dimensional problem and have achieved good results. However, this type of method still heavily relies on the expression capability of the underlying features. In dense crowds, the motion features extracted by current methods often contain a lot of noise, which greatly affects the final result. Additionally, the testing speed and dictionary updates~\cite{zhao2011online} of the sparse coding framework are still critical issues that need to be tackled.
%3方法概述
\section{OVERVIEW OF OUR METHOD}
In this paper, we introduce a method that integrates sparse particle tracking, a repulsive force model and sparse coding to detect and locate abnormal behavior in various crowd videos. Figure~\ref{fig:1} summarizes the main steps of our method.

To avoid tracking all objects, especially in highly dense crowds, a holistic approach is adopted to analyze crowd videos incorporating a grid particle advection method similar to~\cite{mehran2009abnormal,raghavendra2011optimizing}. The tracked particles, which have more stable trajectories, are extracted by a deep optical flow~\cite{weinzaepfel2013deepflow} and a strategy of selecting particle. The repulsive force flow matrix is then obtained by calculating the repulsive forces between the particles. The repulsive force describes the group behavior as a combination of the rejected interactions between individuals, so abnormal behaviors in crowds can be regarded as a consequence of abnormal rejected interactions.

Since the changes in repulsive forces in a crowd scene can instantly reflect the behavior of a crowd, the force flow matrix computed above can be used as the underlying characteristic to describe crowd movement. Our method uses this matrix to extract visual words and constructs different dictionaries by sparse coding. Abnormal events can then be detected and located by sparse reconstruction. Finally, to solve the problems of degradation and concept drift in dictionaries, we propose an unsupervised learning and updating algorithm for the group dictionary model based on sparse reconstruction and a group of word pools. More details will be discussed in Section 6.4.

The rest of this paper is organized as follows. Section 4 describes the selection of characteristic particles and Section 5 introduces the repulsive force model for describing crowd movements. In Section 6, the event representation and abnormal detection are described in detail and an updating strategy is proposed. The experimental results of the global and local abnormal detection are reported in Section 7. This paper is concluded in Section 8.
%4 GRID PARTICLE ADVECTION BASED ON DEEPFLOW
\section{GRID PARTICLE ADVECTION BASED ON DEEPFLOW}
Before computing the repulsive forces in a crowd, it is necessary to identify the movement of each individual. However, as demonstrated in ~\cite{ali2008floor}, it is still challenging to track individuals in high-density crowds due to serious occlusions and high similarity between objects. Therefore, in crowded scenes, individual-based methods are not suitable to estimate repulsive forces. Since a crowd can be considered to be made up of granular particles~\cite{ali2008floor}, some researchers~\cite{ali2007lagrangian,mehran2009abnormal} have attempted to place a grid of particles over the image and move these particles with the flow field computed from the optical flow. These particles are then used to represent individuals in the crowd. Since this type of method is insensitive to crowd density, a similar method has been adopted in our paper using DeepFlow~\cite{weinzaepfel2013deepflow} as the underlying optical flow. DeepFlow blends a novel matching algorithm with a varying approach for optical flow computation, which can handle large-scale movement efficiently in a crowd and provide a more stable tracking trajectory.

\begin{figure}[!htb]
\begin{centering}
 \centering
  \includegraphics[width=0.9\columnwidth]{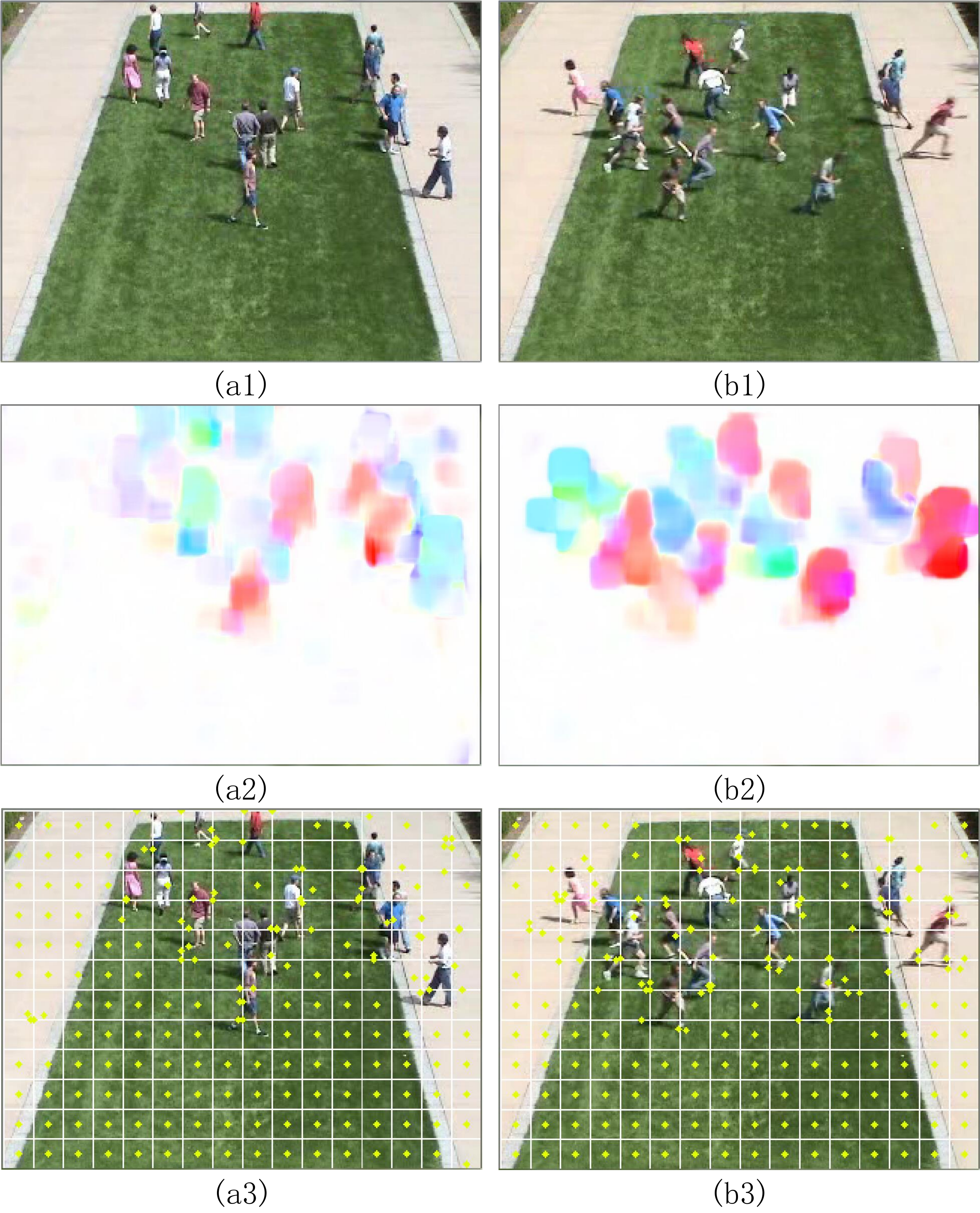}
  \centering
  \caption{Graph of the dense optical flow tracked by DeepFlow and the advection of selected particle. The left column shows normal scenes, and the right column is the abnormal scenes. a1, b1 are the initial images. a2, b2 are the dense optical flow images. a3, b3 are the images of the particle advection.}
\label{fig:2}
\end{centering}
\end{figure}

We have designed a special grid particle advection method based on a dense optical flow computed by DeepFlow, as shown in Figure ~\ref{fig:2}. The optical flow image is firstly divided into regular grid cells. One representative pixel in each cell is then selected to represent the movement trend of that region, which can be seen in Figure ~\ref{fig:2}(a3) and Figure ~\ref{fig:2}(b3). The regular grid is divided using white lines and the yellow dots show the initial tracking points. All chosen pixels represent our sparse tracking pedestrians.

The most important part is to select a pixel as mentioned above, which is named the characteristic particle. For each pixel in the $i-th$ frame at $(x,y)$, its new position $(x^{'},y^{'})$ can be obtained in the $(i+1)-th$ frame by DeepFlow. Algorithm 1 has been designed to choose or update these sparse tracking points.

\begin{algorithm}
\caption{Characteristic Particle Tracking Algorithm}
\KwIn{Consecutive video frames $F_{0}, F_{1}, ..., F_{n}$ tracked by DeepFlow. }
\KwOut{Selected characteristic particles $C_{i}^{j}$ of all input frames}
The initial characteristic particle of each grid cell of frame $F_{0}$ is the central pixel\;
\For{$j=1; j \le n; j++$}
{
\For{$i=0; i \le m; i++$}
{
	For grid cell $G_{i}^{j}$ in $F_{j}$, traverse each pixel and then sort the pixels according to their velocity, $v_{1}>v_{2}>v_{3}>...$\;
	Select $s$ particles with the largest speed in $G_{i}^{j}$, then use the K-means method to calculate one clustering center $C_{i}^{j}$, which is named the characteristic particle\;
}
}
\end{algorithm}

%5 ESTIMATION OF REPULSIVE FORCE IN CROWDS
\section{ESTIMATION OF REPULSIVE FORCE IN CROWDS}
Human crowds can bear a striking resemblance to interacting particle systems, which has prompted many researchers to describe pedestrian dynamics in terms of interaction forces and potential energies. After obtaining the sparse representative particles in each frame, an interaction model can be built between them. In our paper, we modify the method presented in~\cite{karamouzas2014universal} as our underlying feature for crowd movement. This method is based not on the physical separation between pedestrians but on the projected time to a potential future collision, and is therefore fundamentally anticipatory in nature. Remarkably, this simple law can describe human interactions across a wide range of situations, speeds and densities.

\begin{figure}[!htb]
\begin{centering}
 \centering
  \includegraphics[width=0.9\columnwidth]{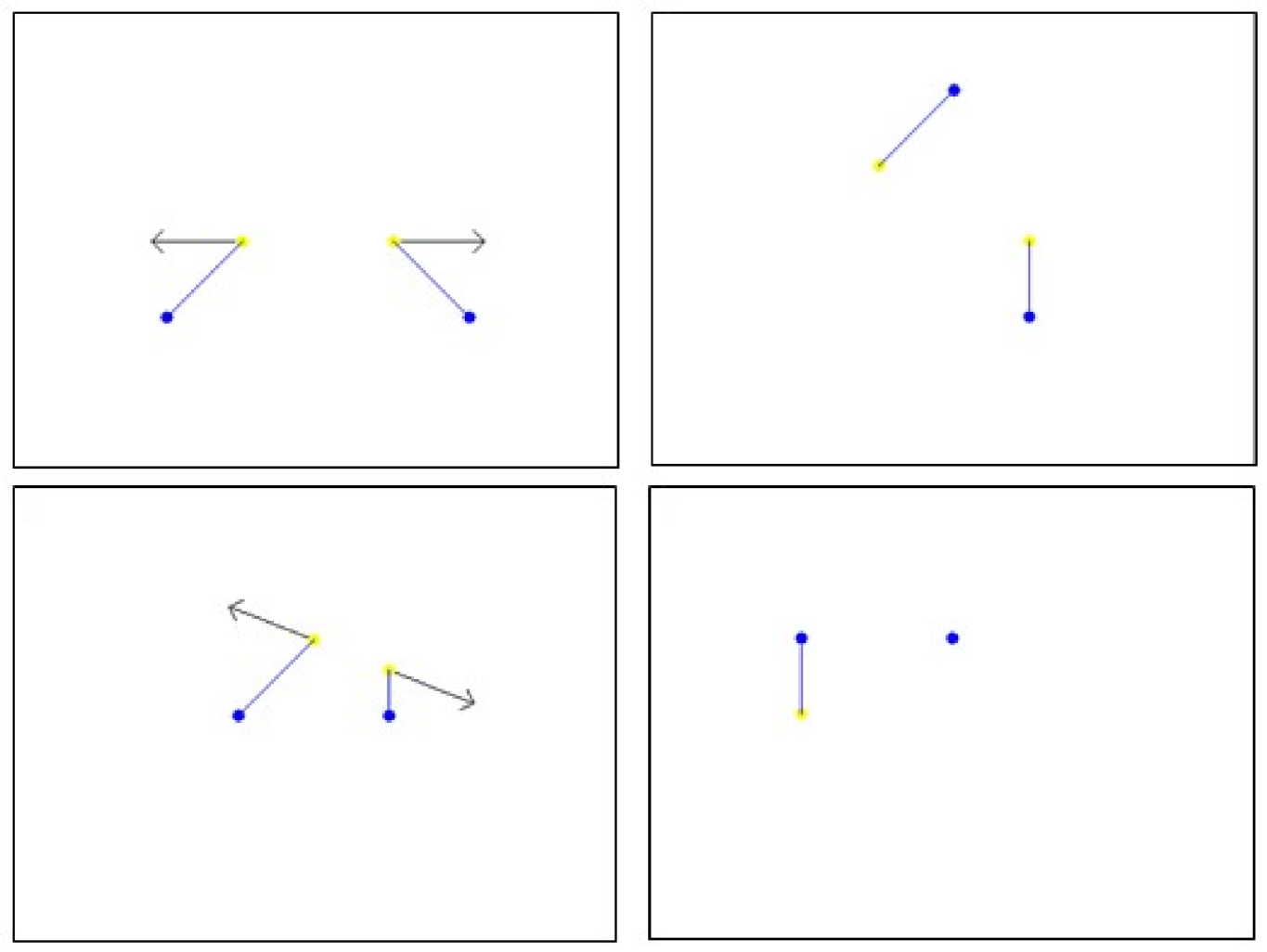}
  \centering
  \caption{Typical examples of repulsive forces between two particles. The blue points are the particle positions in the previous frame and the yellow points are the current positon. The black arrows represent the direction of the repulsive forces. The left column represents relative motion, so a repulsive force exists between the two particles. In the right column, the top figure represents two particles moving away from each other, and the bottom figure represents the scenario where one particle is moving and the other is stationary, thus there is no repulsive force between the two particles.}
  \label{fig:3}
\end{centering}
\end{figure}

According to~\cite{karamouzas2014universal}, the interaction energy between two individuals can be computed using the following equation:\\
\begin{equation}
\label{eqn_example}
E(\tau)=\frac{k}{\tau^2}e^{-\tau/\tau_0}
\end{equation}
$\tau$ is the time-to-collision, $\tau_0$  is the intrinsic range of the unscreened interactions and $k$ is a constant that sets the units of energy. According to the definition of interaction energy, the repulsive force $F$ between two individuals $i$ and $j$ can be implied directly when the pedestrians are interacting by:\\
\begin{equation}
\label{eqn_example}
F=-\bigtriangledown_r(\frac{k}{\tau^2}e^{-\tau/\tau_0})
\end{equation}
where $\bigtriangledown_r$ is the spatial gradient. Some typical simple examples for repulsive forces are shown in Figure ~\ref{fig:3}.

\begin{figure}[t]
\begin{centering}
 \centering
  \includegraphics[width=0.9\columnwidth]{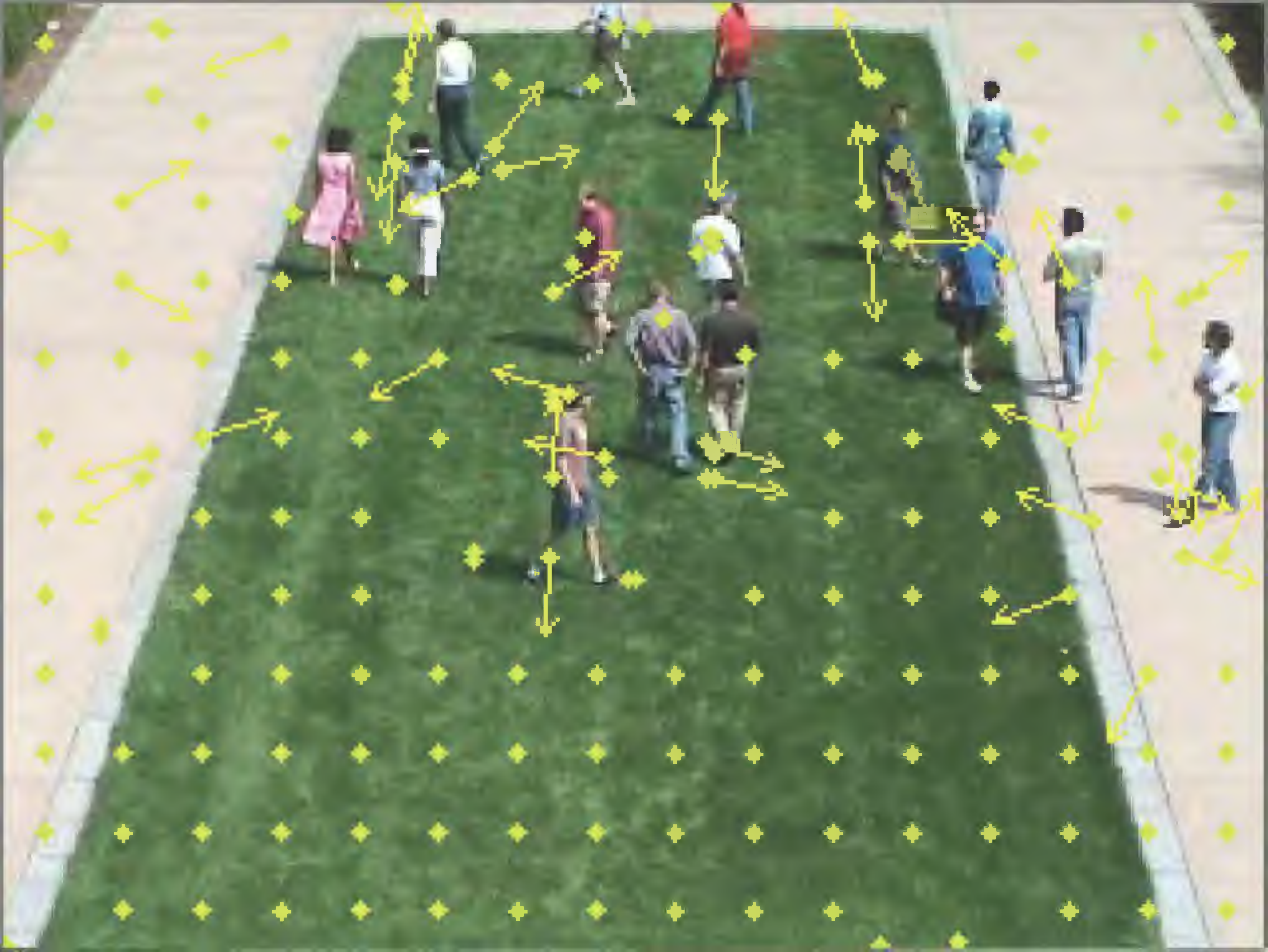}
  \centering
  \caption{Repulsive forces in normal crowd movement. There are a lower number of forces in normal scenes and their value is smaller. The length of the arrows does not represent the size of the forces.}
  \label{fig:4}
\end{centering}
\end{figure}

\begin{figure}[t]
\begin{centering}
 \centering
  \includegraphics[width=0.9\columnwidth]{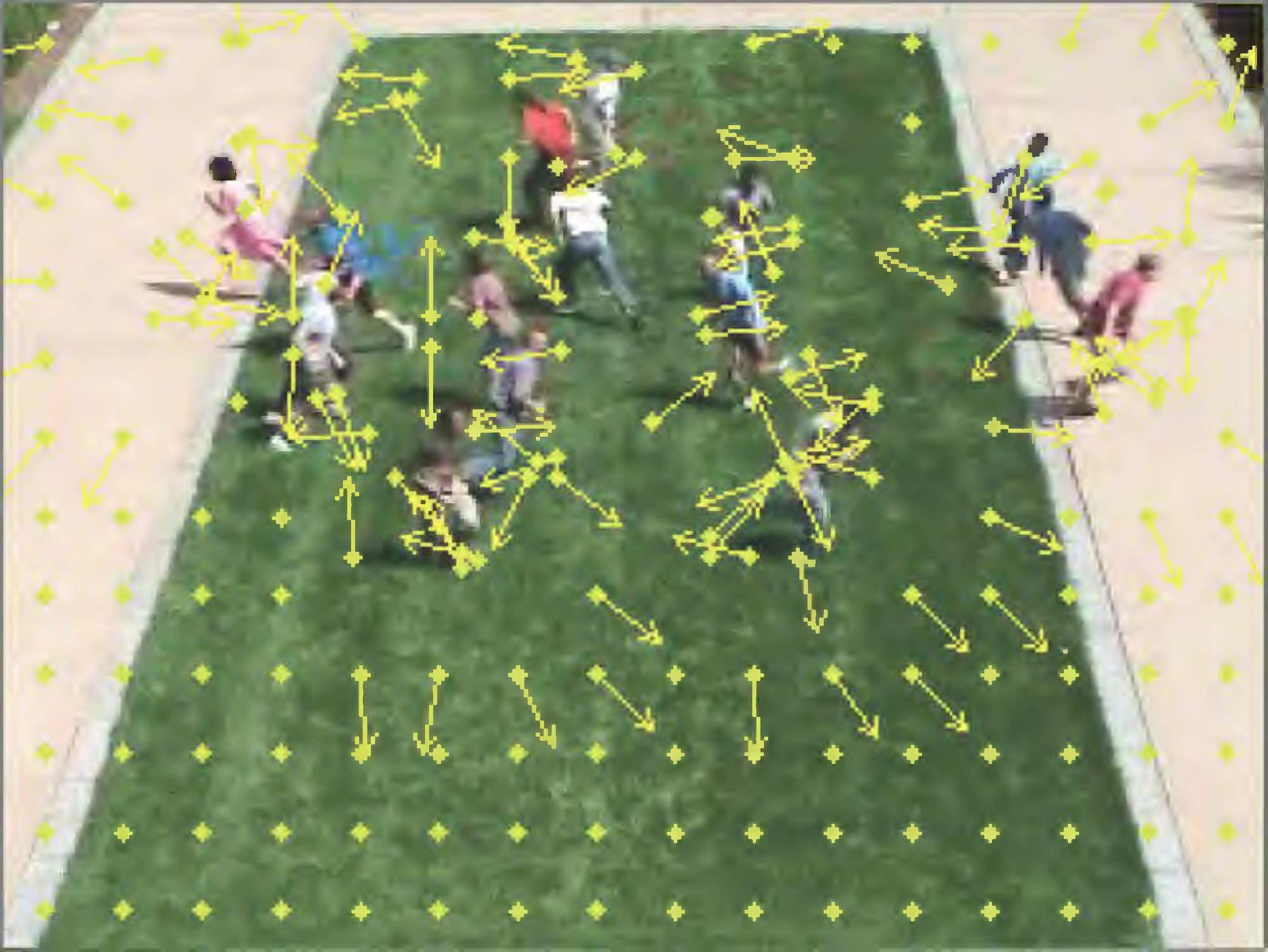}
  \centering
  \caption{Repulsive forces in abnormal crowd movement. There are a higher number of forces in abnormal scenes and their value is bigger. The length of the arrows does not represent the size of the forces.}
  \label{fig:5}
\end{centering}
\end{figure}

In the crowd, each pedestrian will be easily affected by their neighbors. So the resultant force is computed by accumulation of the neighboring forces as follows:\\
\begin{equation}
\label{eqn_example}
F_i=-\sum_{j\in\Omega}F_{ij}
\end{equation}
where $\Omega$ are the neighbors of pedestrian $i$. In our experiments, we found that when $\tau$ is larger than certain threshold, the effect between two neighbors can be ignored. Figure~\ref{fig:4} and Figure~\ref{fig:5} show two typical examples of repulsive forces. The number of forces in normal scenes is less than in abnormal scenes and the value is smaller (the length of the arrows does not represent the size of the forces).

%6 EVENT REPRESENTATION AND ABNORMAL DETECTION
\section{EVENT REPRESENTATION AND ABNORMAL DETECTION}
%6.1 Low-level feature extraction
\subsection{Low-level feature extraction}
The selected low-level visual features have a high influence on the construction of the visual dictionaries. Although the repulsive force calculated in Section 5 can be used as an indicator of potential movement in various crowds, it is still very difficult to distinguish between normal and abnormal behaviors based on discrete force values as presented in~\cite{mehran2009abnormal}. Hence, we adopt a similar force flow method to model the changes in virtual forces among crowds over time.

The training video is firstly partitioned into clips, where each clip consists of $m$ frames. The force flow matrix $F$ is then extracted from each clip. As mentioned in Section 4, each frame in the clip is divided into $n=b\times b$ blocks (b is the number of cells of a row or a column) and the element in $F\in\Re^{m\times n}$ is the 2nd force of the selected particle in each block. Each row $X=\{x_1,...,x_n\}\in\Re^{m\times n}$ is the force change of a single selected point in frame $m$ of the same block, which can be regarded as the initial visual words. The force flow is our underlying characteristic.
%6.2 Initial visual group dictionary generation
\subsection{Initial visual group dictionary generation}
When dealing with high-volume surveillance video data which usually contains complicated scenes, it will become more and more complex to maintain a single visual dictionary. This not only introduces a high computing cost, but also rapidly decreases the expression ability. To avoid this problem, we adopt a group dictionary method, which is similar to that in~\cite{lu2013abnormal}. Each dictionary has a fixed number of trained words. During the process of sparse reconstruction, a suitable dictionary is sought to represent each new word by traversing the group dictionary, rather than judging whether a combination of words exists which can represent this word in a single large dictionary. Additionally, some dominating dictionaries which are obtained firstly can represent most of the normal event features, which can save the search time for suitable dictionaries. To further reduce redundant data and improve detection accuracy, the corresponding visual group dictionaries are trained and updated for different scenes separately. The general concept is to firstly train the initial group dictionary using the first $N$ frames of a new scene and then use the detected events to continuously update the corresponding dictionary~\cite{zhao2011online}.

Specifically, after obtaining the force flow of a scene, the initial visual group dictionary can be estimated as $D=\{D_1,...,D_s\},D_i\in\Re^{m\times n},d<<n$, s is the number of dictionary. To prevent over-fitting, each sub-visual dictionary $D_i$ belongs to a closed, convex and bounded set. The method of training the group dictionary is as follows~\cite{lu2013abnormal}:\\
\begin{equation}
\label{eqn_example}
\begin{split}
\forall{j}\in\{1,...,&n\},t_j=\sum_{i=1}^s\gamma_j^i\{||X_j-D_i\beta_j^i||-\lambda\}\leq0,\\
&s.t.\sum_{i=1}^s{\gamma_j^i=1},\gamma_j^i=\{0,1\}
\end{split}
\end{equation}
where $D_i$ is the $i-th$ visual dictionary, $\beta_j^i$ is the corresponding coefficient set of $X_i$ and $\lambda$ is the reconstruction error upper bound,  $t_j$ is the reconstruction error. $\gamma=\{\gamma_1,...,\gamma_n\},\gamma_j=\{\gamma_j^1,...,\gamma_j^n\}$ and each value of $\gamma_j^i$ indicates whether or not the $i^{th}$ combination $D_i$ is chosen for data $X_j$. The constraints $\sum\gamma_j^i=1$ and $\gamma_j^i=\{0,1\}$ require that only one dictionary is selected.

It is apparent that since the features extracted from different scenes are distinct, there are different dictionary numbers $s$ for different scenes. Actually, during the process of training, the dictionary number S for any particular scene is determined by the reconstruction error upper bound $\lambda$, which is an empirical value.
%6.3 Abnormal event detection
\subsection{Abnormal event detection}
For abnormal event detection, normal events can usually be represented by sparse reconstruction coefficients with a small reconstruction error. In contrast, since abnormal events are extremely different from existing normal events, dense reconstruction coefficients will be produced with a larger reconstruction error. From this observation, abnormal events can be detected by comparing these two factors with a pre-defined threshold.

Since the normal dictionaries are grouped together, the reconstruction error can be computed successively using different dictionaries. If a testing word can be represented by a dictionary, i.e. the reconstruction error of this dictionary is smaller than a pre-defined threshold, the testing word can be regarded as a normal word and the image grid represented by this word has no abnormal events. Meanwhile, this word is given a token to indicate which dictionary it can be represented by. This sign is used to update the subsequent dictionary in section 6.4 If there is no dictionary that can represent a testing word sparsely, it is regarded that an abnormal event has occurred in the region represented by the word.

Specifically, given a new testing word $X$ and a current group dictionary $D=\{D_1,D_2,...,D_s\}$, the reconstruction coefficient $\beta_i,i=1,2,...,s$ can be computed using the above method. $X$ is treated as a normal event if the following criterion is satisfied:\\
\begin{equation}
\label{eqn_example}
R(X,\beta_i,D_i)<\lambda,i=1,2,...,s
\end{equation}
where $R(X,\beta_i,D_i)$ represents the reconstruction error of $X$ and $\lambda$ is the reconstruction error upper bound, whose value is a user-defined experiential threshold that controls the sensitivity of the detection algorithm. As can be seen from algorithm 2, our method of detecting new words can be easily accelerated via parallel processing.
%6.4 Global and local dictionary updating
\subsection{Global and local dictionary updating}
With the increase of new videos and more dynamic changes in crowd scenes, the ability to update online models is critical for stability and wide application of our method. A gradient-descent based method or a simple coding coefficient weighted approach can easily lead to a decrease in the capability of the dictionary representation. In order to solve this problem, a block-coordinate descent~\cite{zhao2011online,lu2013abnormal} is adopted to locally update each dictionary. Additionally, because a group dictionary framework is adopted, the computational cost will be greatly increased if the dictionary is updated each time a new word is detected. The judgment ability of a dictionary will degrade seriously if abnormal words are used to update a dictionary. Hence, only normal words are used to execute the update strategy.
%缺少算法2

\begin{figure}[t]
\begin{centering}
 \centering
  \includegraphics[width=0.9\columnwidth]{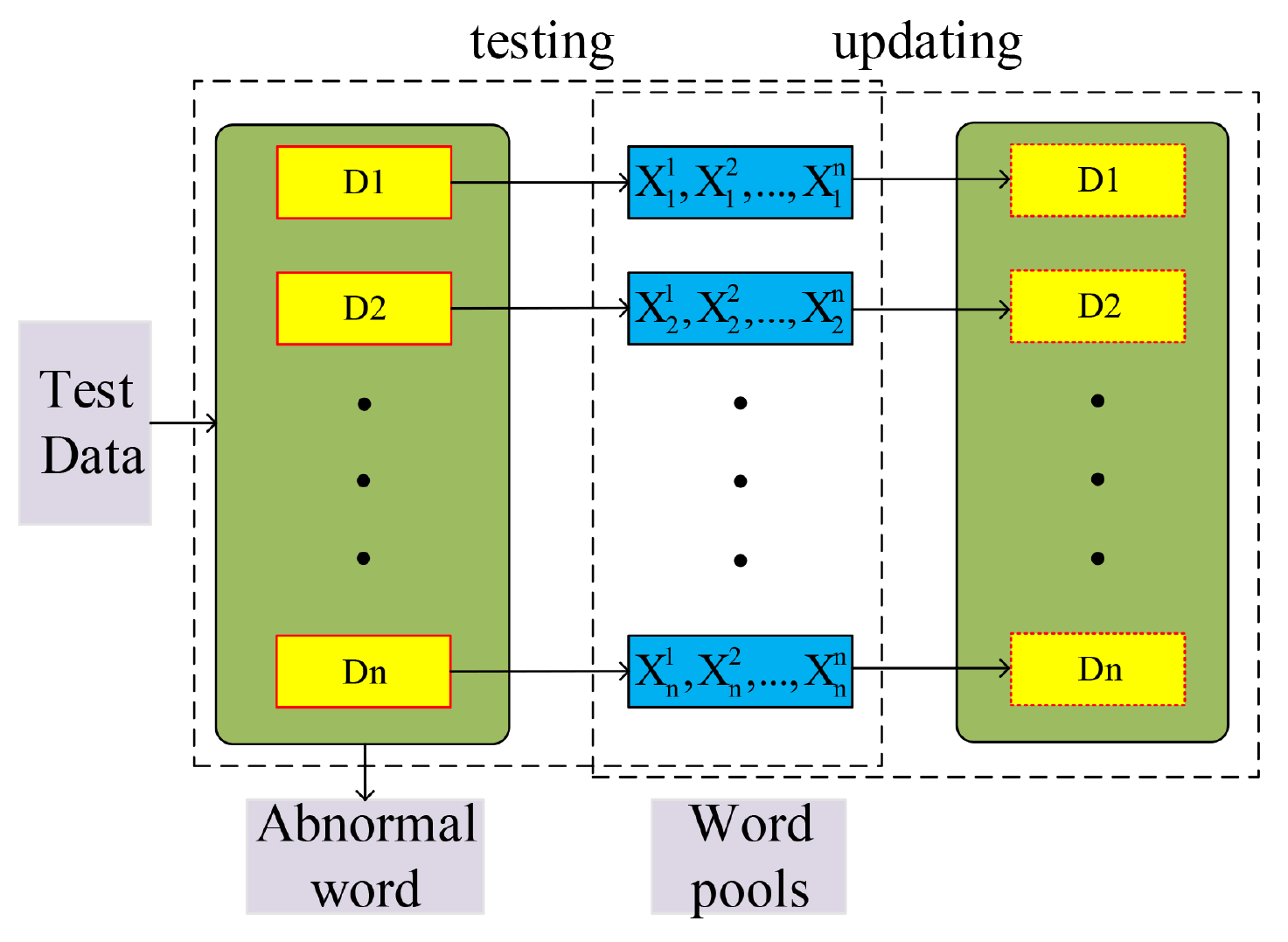}
  \centering
  \caption{Abnormal event detection and local updating for dictionary groups. Each testing word is judged to be a normal event if it can be sparsely represented by one dictionary, and poured into the word pool of the dictionary. Otherwise, it is an abnormal event. The corresponding dictionary is only updated when the number of words in a pool reaches n.}
  \label{fig:6}
\end{centering}
\end{figure}

Since each word has a token to declare which dictionary it belongs to, these normal words can be grouped into different word pools $X_j=\{x_j^1,x_j^2,...,x_j^n\},j=1,2,...,s$ to update their corresponding dictionary separately, as shown in Figure ~\ref{fig:6}.

In order to further reduce the computing cost, the strategy is adopted to perform the updating and detection in parallel. A word pool is used to update the corresponding duplicate dictionary only when the number of words in the pool reaches $n$, until then, the current dictionary is used to detect the video. The value of $n$ is decided for different scenes. Generally, complex scenarios have a smaller $n$ and simple scenarios have a bigger $n$. Once the update is completed, the current dictionary is replaced with the newest dictionary for the next round of detection and updating. Specifically, the block-coordinate descent is used to update the dictionary:\\
\begin{equation}
\label{eqn_example}
D_i^{'}=\prod[D_i-\delta_t \bigtriangledown_{_{D_i}} L(\beta,D_i)]
\end{equation}
where $D_i^{'}$ is the newest dictionary. $L(\beta,D_i)=\sum_{j=1}^{n}||X_j-D_i\beta_j^i||_2^2,j=1,...,n$ indicates that the update is executed only when the number of words in the word pool of $D_i$ reaches $n$. $\beta_j^i=(D_i^{T}D_i)^{-1}D_i^{T}X_j$ and $\delta=1E-4$, $\prod$ denote projecting the basis to a unit column.

Since each dictionary in the group dictionary is updated separately, the situation may arise where one dictionary gradually becomes closer to another one. The intuitive indication is that a new word can be represented by more than one dictionary after a period of time. Once this situation emerges, all of the words in each word pool need to be used to re-train the group dictionary to execute a global update. Since our algorithm firstly conducts local updates and then executes global updates, this guarantees that group dictionaries will change with scenes. Meanwhile, many concrete implementations of our algorithm can use parallel execution, thus ensuring holistic detection efficiency. Algorithm~\ref{alg:2} and Algorithm~\ref{alg:3} describe the specific process of event detection for global and local updating.

\begin{algorithm}
\caption{Abnormal Event Detection.}
\label{alg:2}
\KwIn{new input testing video\;
reconstruction error upper bound $\lambda$\;
initial visual group dictionaries $D={D_{1},D_{2},...,D_{S}}$ trained with first $N$ frames in normal video.}
\KwOut{whether and where abnormal behavior happens in testing video.}
\For{each word $x$ extracted from testing video}
{
	\For{$i=1;i \leq s-1; i++$}
	{
		compute $\beta_{i},\beta_{i+1}$ with Eq.(4) and the latest		$D_{i},D_{i+1}$\;
	\eIf{$R(x,\beta_{i},D_{i})< \lambda $}
{
\eIf{$R(x,\beta_{i+1},D_{i+1})< \lambda$}
	{use all the words in word pools to execute global dictionary updating\;
	return normal event.}
	{put $x$ into the word pool of $D_{i}$\;
	call Algorithm 3 to execute local dictionary updating\;
	return normal event.}
}
{
	return abnormal event.}

	}
}
\end{algorithm}

\begin{algorithm}
\caption{Local Visual Dictionary Updating.}
\label{alg:3}
\KwIn{dictionary $D_{i}$ with its word pool $X_{i}=\{x_{i}^{1},{x_{i}^{2},...\}}, i=1,2,...,m$.}
\KwOut{newest $D_{i}$.}
\If{the number of word in word pool $X_{i}>n$}
{
	construct an empty dictionary $D_{i}^{'}$\;
	swap($D_i$, $D_{i}^{'}$)\;
	update $D_{i}^{'}$ with Eq.(5)\;
	swap($D_i$, $D_{i}^{'}$)\;
}
\end{algorithm}

%缺少算法3
%6.5 Localization of Abnormalities
\subsection{Localization of Abnormalities}
After obtaining the newest group dictionary, we can implement sparse reconstruction for the input testing video. As described previously, since our approach is based on a regular image grid, the corresponding reconstruction error can be easily computed for each grid. When the error is larger than a set threshold, the event can be regarded as an abnormal event in that region. Compared with other methods, our approach is more intuitive and simple. More details are given in Section 7.
%7 EXPERIMENTS AND DISCUSSION
\section{EXPERIMENTS AND DISCUSSION}
In this section, two publicly available datasets and one self-organizing dataset are adopted to validate the effectiveness of our method under different scenarios. As in~\cite{mehran2009abnormal,cong2013abnormal}, the UMN dataset is used to verify the ability of the algorithm to detect global abnormal events. As in~\cite{zhu2014sparse,li2014anomaly}, the UCSD dataset is applied to detect local abnormal events. Finally, the self-organizing dataset is used to validate the effectiveness and generality of our method for realistic crowd scenes. Unlike most conventional methods which obtain a fixed dictionary based on a training video, and then detect abnormal events from new crowd video segments, our method mainly focuses on online abnormal detection and location for a single scene. During the initialization stage, a period of normal video is used to train the dictionary group~\cite{lu2013abnormal}, and then the detected events are used to update the corresponding dictionary continuously using an updating strategy~\cite{zhao2011online}. This can not only greatly reduce the redundancy of the dictionary and increase the speed of detecting abnormal events, but also effectively guarantees that the dictionary ability will not be degraded over time.
%7.1 The UMN Dataset
\subsection{The UMN Dataset}
The UMN dataset contains three different scenes and 11 segments of video sequence. The total number of frames is 7740 with a $320\times240$ resolution. At the beginning of each video sequence, the crowd behavior is normal. After a period of time, the crowd begins to flee with panic.

Training and testing are performed separately for each of the three scenarios. More specifically, for each scenario, 500 normal frames are used to train the initial dictionary group, and the remainder of the data is used to update and test the initial trained dictionary group. Firstly, the video stream, which is tracked and processed by dense optical flow, is divided into $20\times20$ blocks without overlaps to calculate the repulsive force in each image block, as shown in Figure ~\ref{fig:1}. According to practical experience, when handling this type of scenario, the formula parameters should be set to $k=1.5,t_0=3s$ and the reconstruction error upper bound should be set to $\lambda=0.08$. During the training process, since the interaction force is computed based on a normal crowd video, the value of  $F$ is small and stable. If the value of $F$ of the testing frame increases abruptly, this indicates that a particular abnormal crowd event has appeared. After obtaining the repulsive force of the representative particle of every frame in the image block, the change in force between successive frames in the same block can be regarded as the movement of people in the video. In order to construct visual words, the force flow is divided into a force flow matrix section by section with $T=30$ frames. Each row of the matrix represents one visual word. For the training phase, there are approximately 3100 visual words. Approximately 42 dictionary groups are then obtained by sparse reconstruction training~\cite{zhao2011online,cong2013abnormal,lu2013abnormal}.
\begin{figure}[t]
\begin{centering}
 \centering
  \includegraphics[width=0.9\columnwidth]{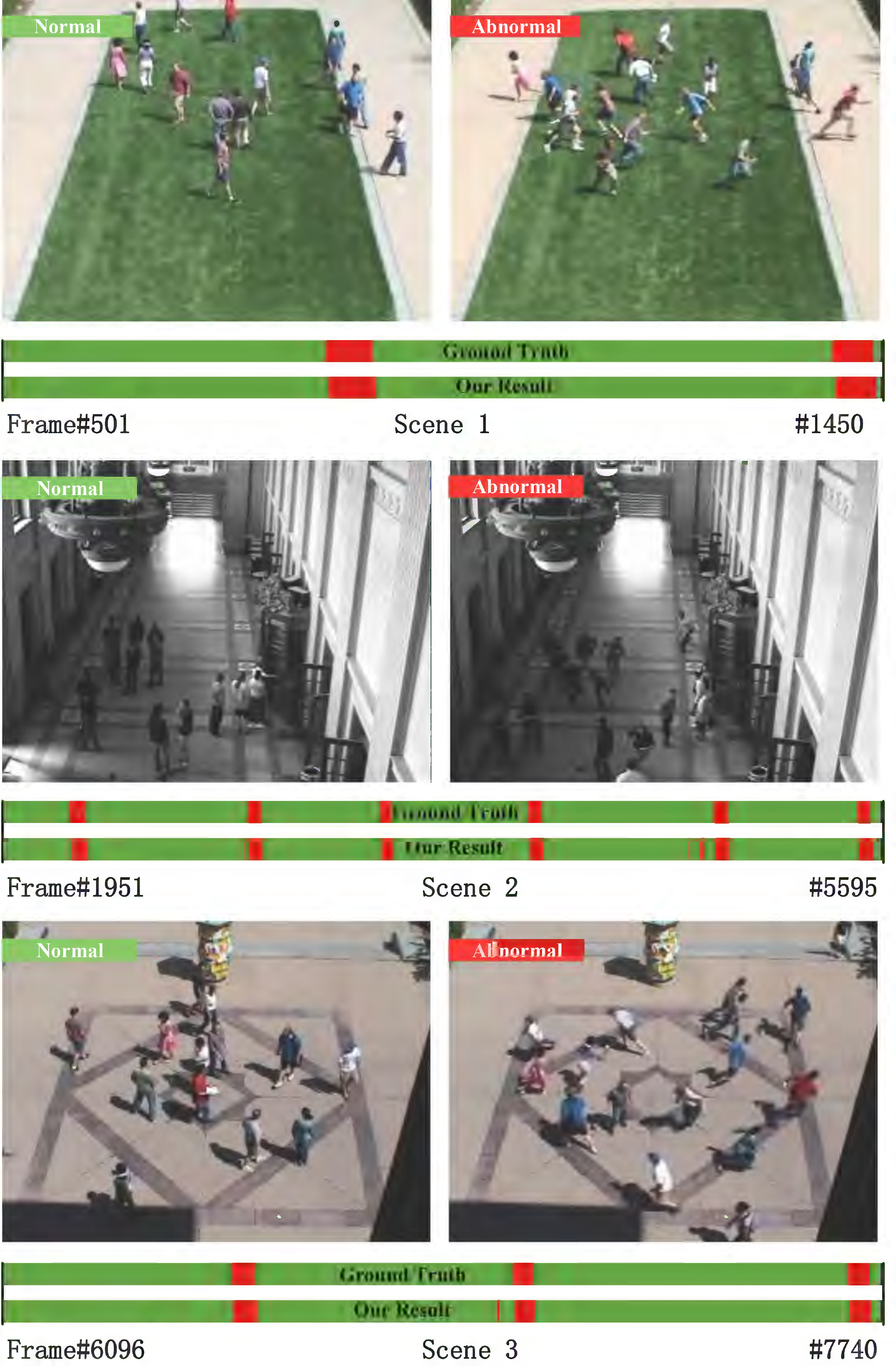}
  \centering
  \caption{The qualitative results of abnormal behavior detection for the three scenarios of the UMN dataset. The column on the left shows normal scene conditions, the column on the right shows scenes with anomalies. The first color bar represents the ground truth and the second color bar denotes our test results. Green is normal and, red denotes exceptions.}
  \label{fig:7}
\end{centering}
\end{figure}

\begin{figure}[t]
\begin{centering}
 \centering
  \includegraphics[width=0.9\columnwidth]{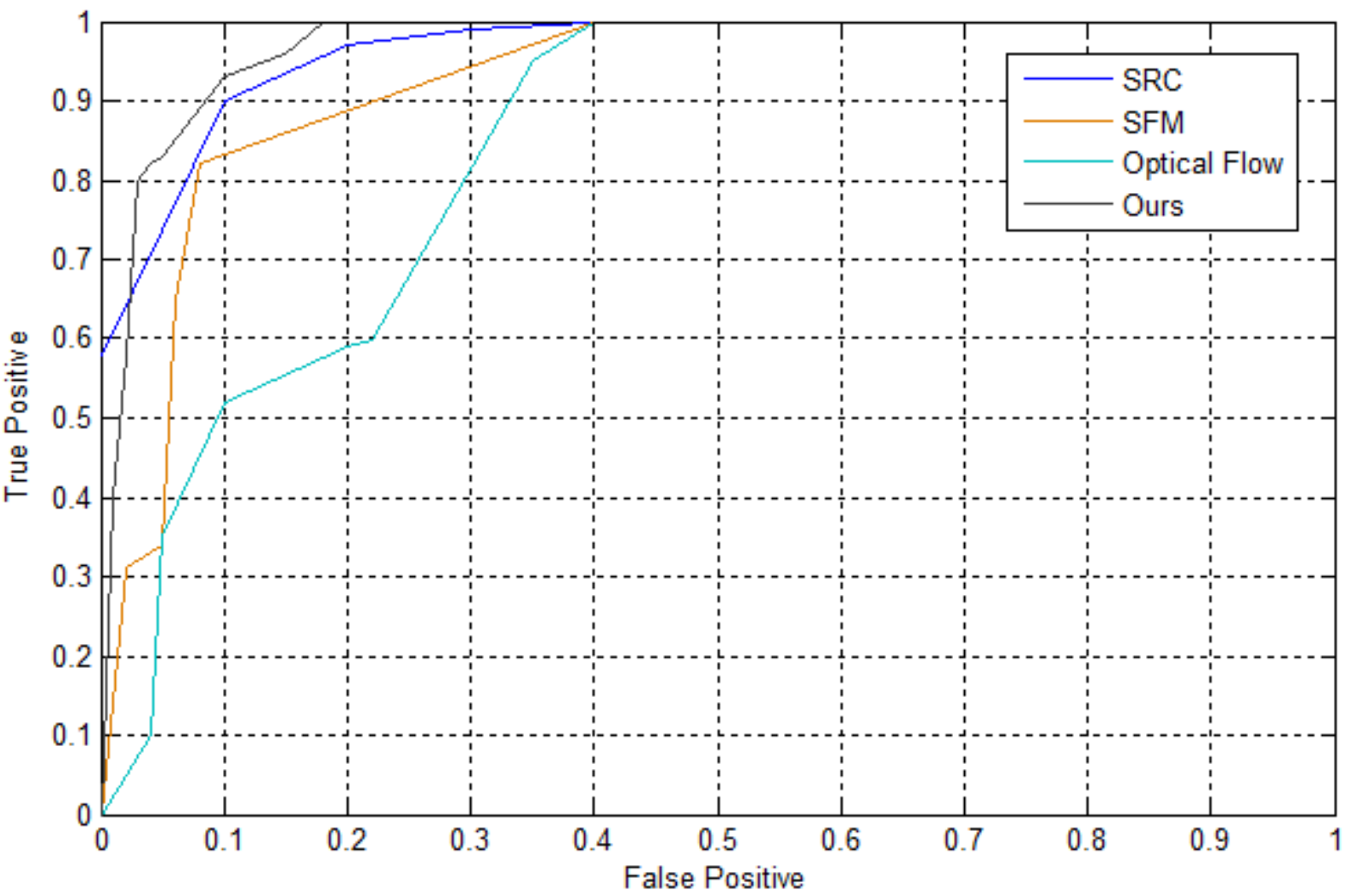}
  \centering
  \caption{Comparison analysis of performance for abnormal locations on the UMN dataset.}
  \label{fig:8}
\end{centering}
\end{figure}

For the testing phase, the same method is adopted to obtain visual words, and the coefficient weight vector is then computed using the above sparse reconstruction. Using these weights, we attempt to judge whether abnormal behavior has appeared at the visual word location and whether the word has been added to a word pool which can sparsely represent it. A dictionary is updated when the number of words $n=4000$ in its corresponding word pool. Figure ~\ref{fig:7} shows the global anomaly detection results. When an abnormal event occurs, the weight coefficients are large and much denser than for normal events by several orders of magnitude. If the visual word is normal, it is used to update the dictionary group. Furthermore, during the force flow construction process, the spatial and temporal information are fully considered, and the group dictionary method real-time updating are adopted during detection, so our method can guarantee a lower false alarm rate, as shown in Figure ~\ref{fig:8}.

%7.2 UCSD Dataset
\subsection{UCSD Dataset}
\begin{figure}[t]
\begin{centering}
 \centering
  \includegraphics[width=0.9\columnwidth]{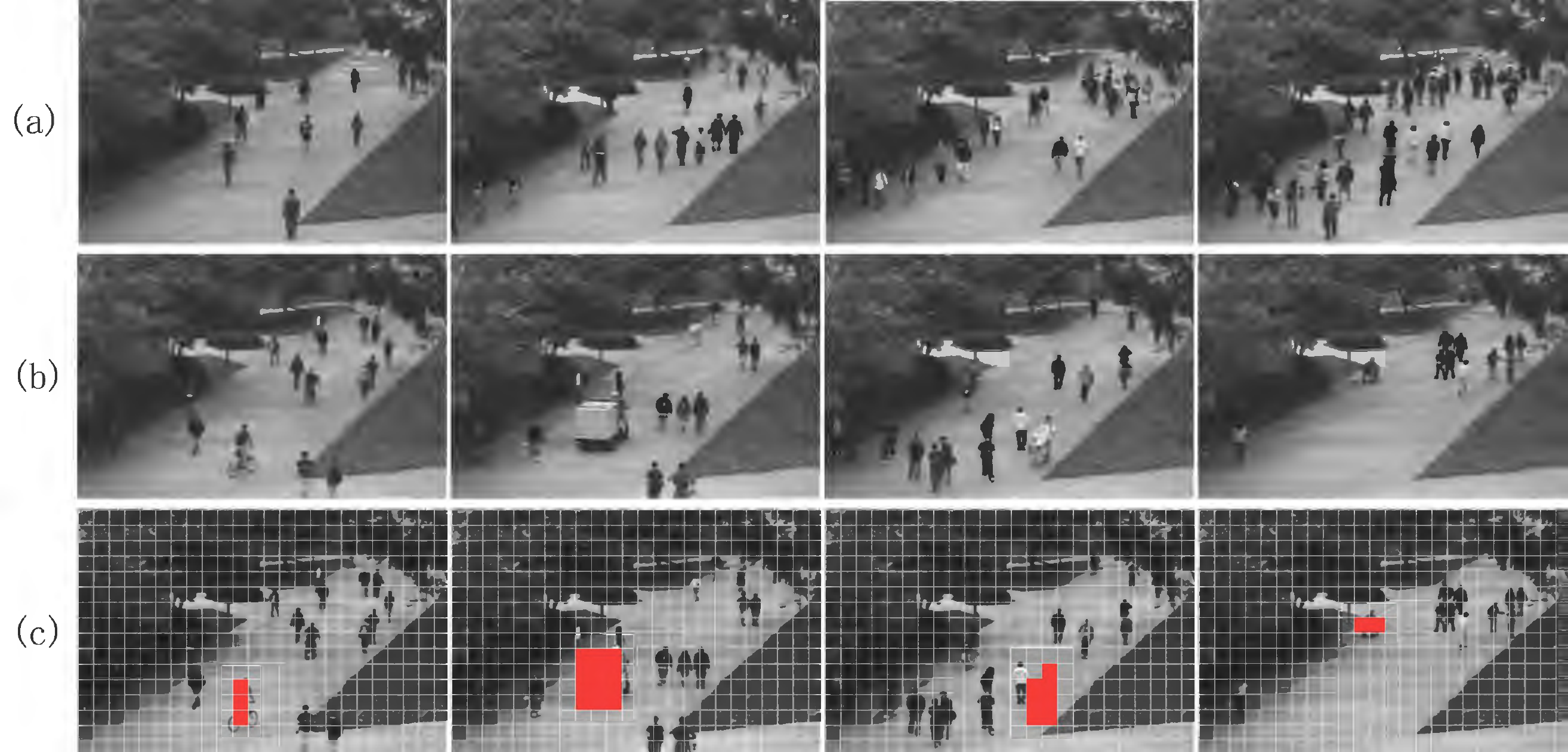}
  \centering
  \caption{Some UCSD dataset scenes. (a) shows normal scenes from sparse to dense, (b) shows some scenes with the appearance of a biker, small carts etc. (c) illustrates that our method can locate the abnormal position within the scenes.}
 \label{fig:9}
\end{centering}
\end{figure}

The UCSD dataset is collected from a fixed camera on the sidewalk. It includes two sub-datasets named Ped1 and Ped2, which correspond to different scenarios. The crowd density of this dataset ranges from sparse to dense, and there also exists serious occlusion between individuals in the crowd. In the normal scenario, the video only contains pedestrians. Abnormal events occur when non-pedestrian objects appear on the sidewalk or the pedestrians perform abnormal movements. In this dataset, abnormal events include bikers, skaters, small carts, people who are walking across a walkway and so on. Since there is no pixel-level ground truth in Ped2, we mainly use Ped1 to verify the performance of the proposed method to locate anomalies. The Ped1 sub-dataset includes 34 segments of normal training video and 36 segments of testing video. Each segment consists of approximately 200 frames with a resolution of $158\times238$. Figure ~\ref{fig:9}(a) and (b) show some sample frames of Ped1.

To train Ped1, the same method discussed in Section 7.1 is applied to train the group dictionary. Based on practical experience, the video stream is divided into $10\times10$ non-overlapping blocks to calculate the repulsive force. The reconstruction error upper bound is $\lambda=0.06$ and the other parameters are set as follows: $k=1.5,t_0=2s,T=20$ frames. The updating strategy is executed to update the dictionary when the number of words $n=2000$ in its corresponding word pool.

\begin{table}[t]
\begin{tabular}{p{2.5cm}<{\centering} p{2.5cm}<{\centering} p{2.5cm}<{\centering}}
\hline
\hline
Methods & EER(\%) & RD(\%) \\
\hline
SFM 	&	31	& 	21	\\
MPPCA	&	40	&	18	\\	
MDT		&	25	&	45	\\
Spare+LSDS	&	19	&	46	\\
HOS         &   27  &   75  \\
Ours		&	14	&	77	\\
\hline
\end{tabular}
\caption{Statistical results of UCSD Ped1 dataset. Quantitative comparison of our method with [30] and HOS~\cite{Kaltsa2015anomaly}. EER is the equal error rate and RD is the rate of detection.}
\label{tb:1}
\end{table}

\begin{figure}[t]
\begin{centering}
 \centering
  \includegraphics[width=0.9\columnwidth]{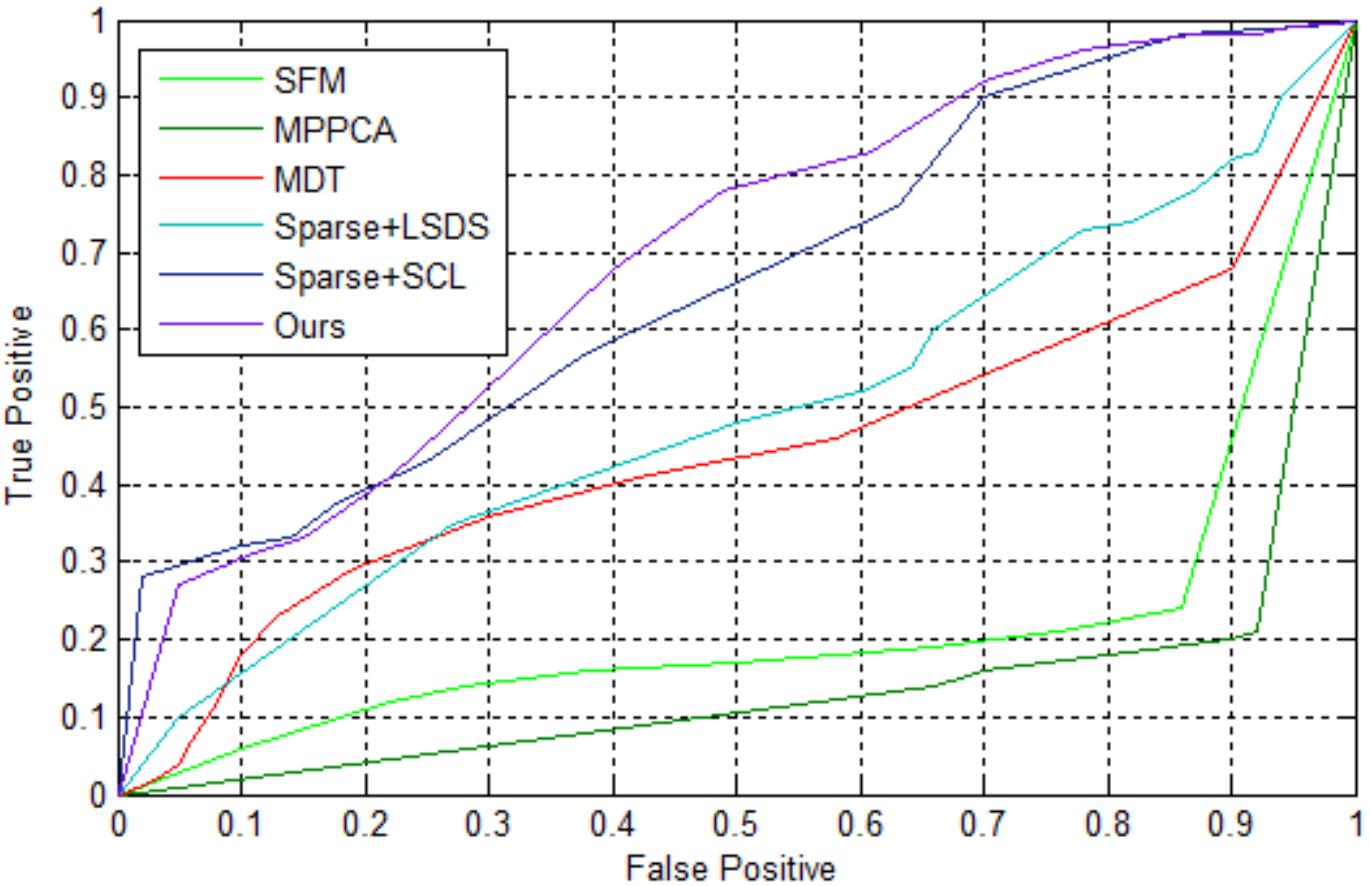}
  \centering
  \caption{A pixel-level comparison analysis of the performance at the abnormal locations on the Ped1 sub-dataset. Method abbreviations: SFM~\cite{mehran2009abnormal}, MPPCA~\cite{kim2009observe}, MDT~\cite{mahadevan2010anomaly}, Sparse + SCL~\cite{lu2013abnormal} and Sparse + LSDS~\cite{cong2013abnormal}.}
\label{fig:10}
\end{centering}
\end{figure}

\begin{figure}[t]
\begin{centering}
 \centering
  \includegraphics[width=0.9\columnwidth]{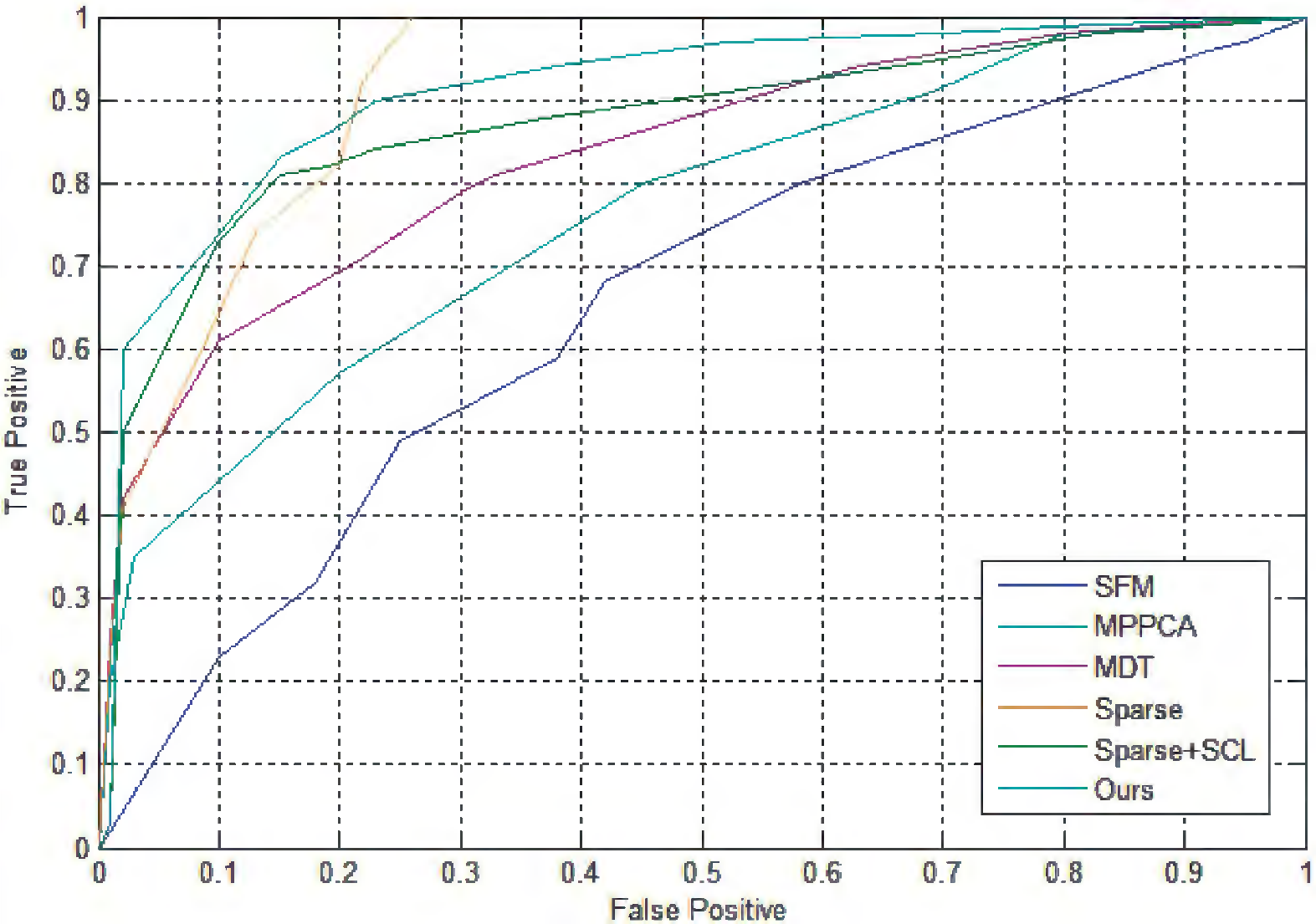}
  \centering
  \caption{A frame-level comparison analysis of the performance at the abnormal locations on the Ped1 sub-dataset. Method abbreviations: SFM~\cite{mehran2009abnormal}, MPPCA~\cite{kim2009observe}, MDT~\cite{mahadevan2010anomaly}, Sparse + SCL~\cite{lu2013abnormal}.}
\label{fig:11}
\end{centering}
\end{figure}

The detection results shown in Figure ~\ref{fig:9}(c) indicate that the proposed method can accurately locate abnormalities. For frame level detection, a frame is considered to be a successful detection if it contains at least one abnormal pixel. In our experiment, if a frame contains one or more abnormal grid cells, it is labeled as an abnormal frame. For the pixel level evaluation, the method in~\cite{mahadevan2010anomaly} is followed, and if more than 40\% of truly anomalous pixels are detected, the corresponding frame is considered as being correctly detected. The ROC curves in Figure $~\ref{fig:10} \& ~\ref{fig:11}$ show the comparison results at a frame level and a pixel level respectively, using the other methods. Table~\ref{tb:1} shows the superiority of our approach, both in terms of the equal error rate (EER) and the detection rate (RD). Our method achieves satisfactory performance, because our method trains a group of visual dictionaries for a single scene and updates the dictionaries in real time.

%7.3 Web Dataset
\subsection{Web Dataset}
\begin{figure}[t]
\begin{centering}
 \centering
  \includegraphics[width=0.9\columnwidth]{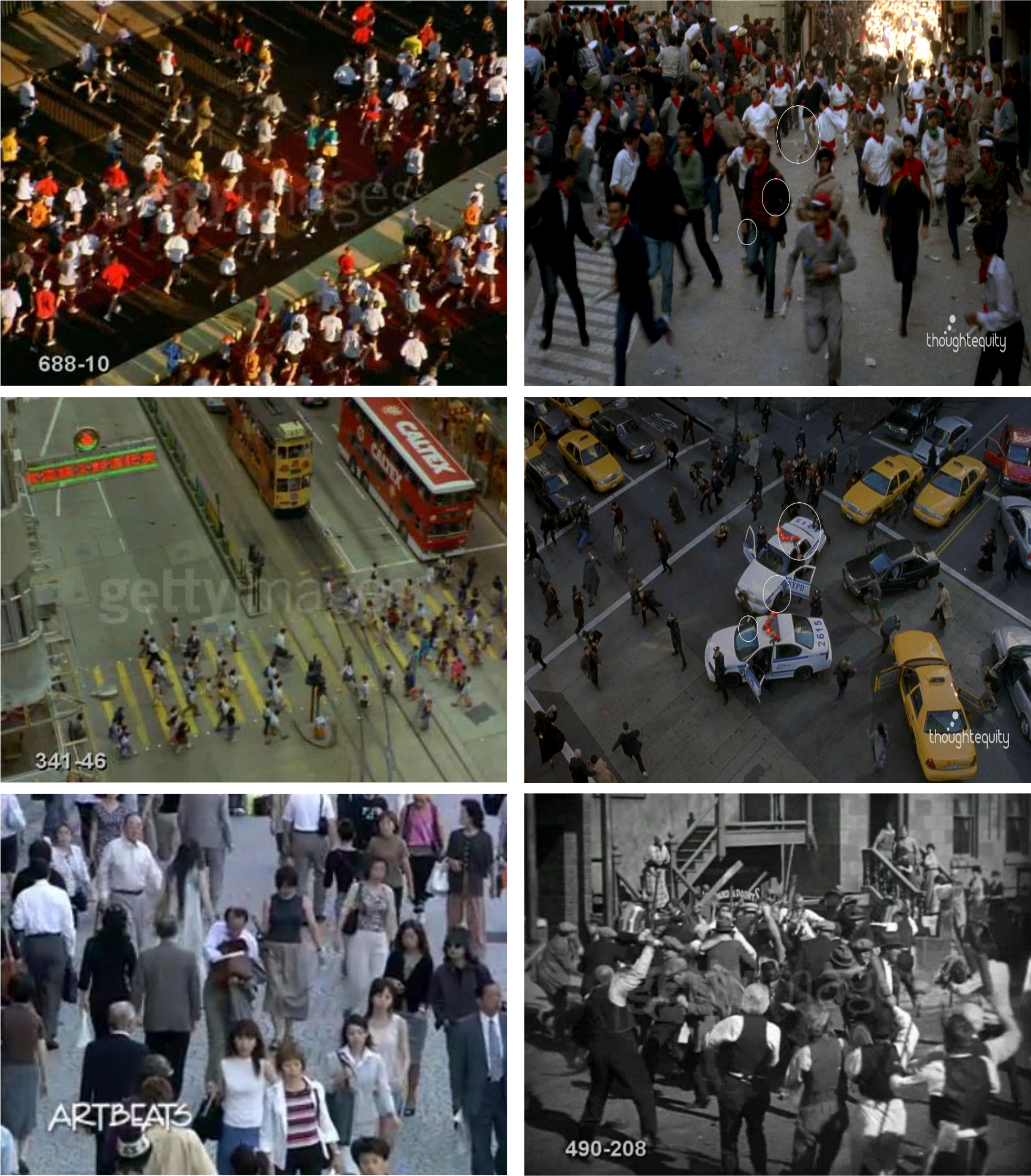}
  \centering
  \caption{Some example scenes from the Web dataset. The left column shows normal samples and the right column shows abnormal samples.}
  \label{fig:12}
\end{centering}
\end{figure}

To further verify the suitability of our proposed method for different scenarios, and also evaluate the effectiveness of this method in actual scenes, this method is applied to the Web Dataset~\cite{mehran2009abnormal} which is closer to a realistic scene. This dataset contains twelve normal crowd scenes, such as pedestrians walking and a marathon, and contains eight abnormal scenes, such as fighting, escapes and so on. Figure ~\ref{fig:12} shows some sample frames of the dataset.

\begin{figure}[t]
\begin{centering}
 \centering
  \includegraphics[width=0.9\columnwidth]{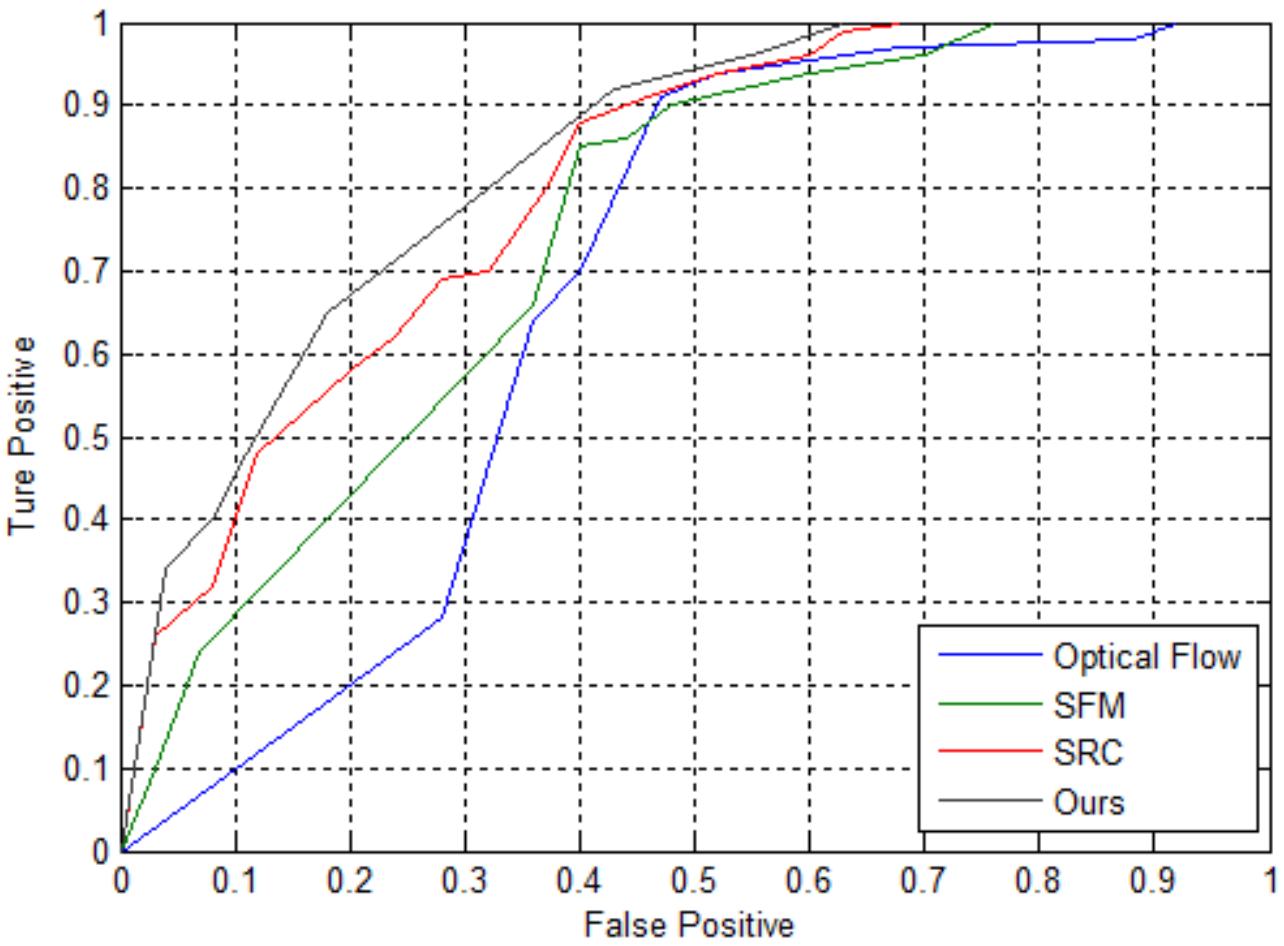}
  \centering
  \caption{The ROCs of abnormal behavior detection in the web dataset.}
  \label{fig:13}
\end{centering}
\end{figure}

Since each video segment in this dataset is short, some of them are used to train the group dictionary and the others are used for testing. In detail, ten normal segments are randomly selected for training and the two remaining normal video segments and eight abnormal video segments are used for testing. Based on practical experience, the video stream is divided into $20\times20$ blocks and the other parameters are set as: $\lambda=0.04,k=1.5,t_0=2s,T=20,n=2000$. As in~\cite{mehran2009abnormal}, to achieve accurate and stable results, the segments are randomly selected ten times to verify the training and the ROC curve is then obtained by averaging the experimental results ten times. On the one hand, this method fully considers spatial-temporal information and regular population movement~\cite{karamouzas2014universal}, on the other hand, it detects abnormal regions based on a fixed grid. Therefore, the final result is better than SFM and SRC, as shown in Figure ~\ref{fig:13}.

%8 CONCLUSION
\section{CONCLUSION}
This paper proposes a method that combines a repulsive force model and sparse reconstruction to detect and locate abnormal events in crowded scenes. Compared with other methods, our main contribution is that we have achieved more stable underlying feature extraction and more efficient and accurate detection and localization of abnormal events. In addition, for a single scene, the group dictionary and online update method adopted by this paper effectively solves the problem of inadequate representation capability and degeneration of the dictionary. Our future work is to further improve our algorithm speed and adaptability to extend the usage range of our algorithm.

%\appendices

%\section{}

%\ifCLASSOPTIONcompsoc
%\else
%\fi


\ifCLASSOPTIONcaptionsoff
  \newpage
\fi

\bibliographystyle{IEEEtran}
\bibliography{mybiblunwen}

\begin{IEEEbiography}[{\includegraphics[width=1in,height=1.25in,clip,keepaspectratio]{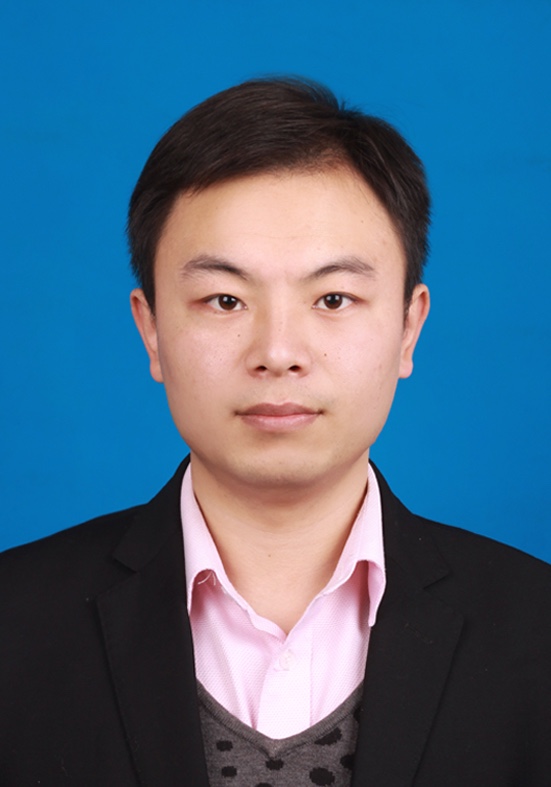}}]{Pei Lv}
is an assistant professor in Center for Interdisciplinary Information Science Research, School of Information Engineering, Zhengzhou University, China. His research interests include video analysis and crowd simulation. He received his Ph.D in 2013 from the State Key Lab of CAD\&CG, Zhejiang University, China.
\end{IEEEbiography}

\begin{IEEEbiography}[{\includegraphics[width=1in,height=1.25in,clip,keepaspectratio]{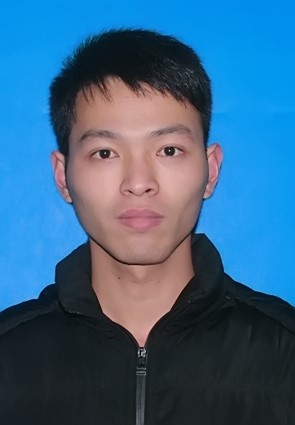}}]{Shunhua Liu}
is a graduate student in the School of Information Engineering of Zhengzhou University, China. He received the bachelor's degree in computer science and technology from Zhengzhou University. His research interests primarily include computer vision, machine learning, and pattern recognition techniques for smart video surveillance systems, with specific focus on human behavior analysis, visual tracking in crowded scenarios.
\end{IEEEbiography}

\begin{IEEEbiography}[{\includegraphics[width=1in,height=1.25in,clip,keepaspectratio]{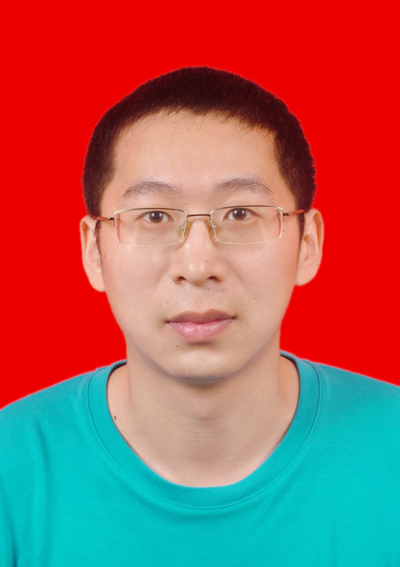}}]{Mingliang Xu}
is an associate professor in the School of Information Engineering of Zhengzhou University, China, and currently is the director of CIISR ( Center for Interdisciplinary Information Science Research), and the general secretary of ACM SIGAI China. His research interests include virtual reality and artificial intelligence. Xu got his Ph.D. degree in computer science and technology from the State Key Lab of CAD\&CG at Zhejiang University.
\end{IEEEbiography}

%\begin{IEEEbiography}[{\includegraphics[width=1in,height=1.25in,clip,keepaspectratio]{LinNie.jpg}}]{Meng Wang}
%is an Assistance Researcher with the Center for Shared Experimental Education, Sun Yat-Sen University (SYSU), China. She received her B.S. degree from the Beijing Institute of Technology (BIT), China in 2004 and her M.S. degree from the Department of Statistics, University of California, Los Angeles (UCLA). Her research focuses on data mining and artificial intelligence.
%\end{IEEEbiography}

\begin{IEEEbiography}[{\includegraphics[width=1in,height=1.25in,clip]{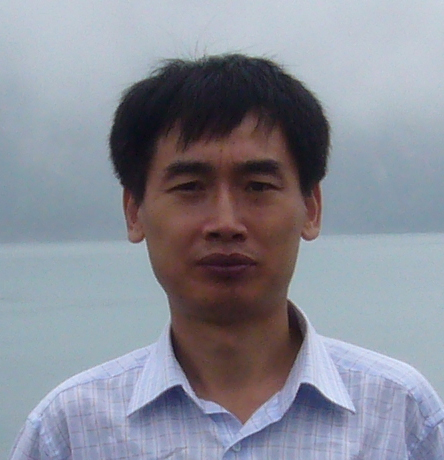}}]{Bing Zhou}
is currently a professor at the School of Information Engineering, Zhengzhou University, Henan, China. He received the B.S. and M.S. degrees from Xian Jiao Tong University in 1986 and 1989, respectively,and the Ph.D. degree in Beihang University in 2003, all in computer science. His research interests cover video processing and understanding, surveillance, computer vision, multimedia applications.
\end{IEEEbiography}
\end{document}